\documentclass[11pt,a4paper]{article}
\usepackage[hyperref]{emnlp2020}
\usepackage{times}
\usepackage{latexsym}

\usepackage{amsmath,graphicx}
\usepackage{soul}
\usepackage{color}
\usepackage{mathrsfs}
\usepackage{algpseudocode}
\usepackage{multirow}
\usepackage{subfigure}
\usepackage{pifont}
\usepackage{bm}
\usepackage{booktabs}
\usepackage{tikz}
\usepackage{amssymb}
\usepackage{textcomp}
\usepackage{colortbl}
\usepackage{bibentry}
\usepackage{ulem}
\usepackage{float}
\usepackage{arydshln}
\usepackage{threeparttable}

\usepackage{flexisym}
\usepackage{url}
\usepackage{hyperref}
\hypersetup{colorlinks,linkcolor=blue}
\usepackage{tabularx}
\usepackage{scrextend}

\colorlet{soulred}{red!50}

\colorlet{soulbleu}{cyan!20}

\colorlet{soulgreen}{green!20}

\colorlet{soulyellow}{yellow!40}

\colorlet{soulorange}{orange!30}

\colorlet{soulpurple}{blue!50}

\usepackage{microtype}

\aclfinalcopy 

\title{Dimsum @LaySumm 20: BART-based Approach for Scientific Document Summarization}

\author{Tiezheng Yu$^{1}$, Dan Su$^{1,2}$,  Wenliang Dai$^{1}$, Pascale Fung$^{1,2}$ \\
$^{1}$Center for Artificial Intelligence Research (CAiRE)\\
The Hong Kong University of Science and Technology, Clear Water Bay, Hong Kong\\
$^{2}$EMOS Technologies Inc. \\
\tt \{tyuah,dsu,wdaiai\}@connect.ust.hk, \\ 
\tt pascale@ece.ust.hk}

\date{}

\begin{document}
\maketitle
\begin{abstract}
Lay summarization aims to generate lay summaries of scientific papers automatically. It is an essential task that can increase the relevance of science for all of society. In this paper, we build a lay summary generation system based on the BART model. We leverage sentence labels as extra supervision signals to improve the performance of lay summarization. In the CL-LaySumm 2020 shared task, our model achieves 46.00\% Rouge1-F1 score.
\end{abstract}

\section{Introduction}
Nowadays, researchers have been increasingly tasked by funders and publishers to outline their research for the public by writing a lay summary. Therefore, it is essential to automatically generate lay summaries to reduce the workload for researchers as well as build a bridge between the public and science. Previous studies have investigated scientific article summarization especially for papers \cite{cohan2018discourse, lev2019talksumm, yasunaga2019scisummnet}. However, less work has been done to generate lay summaries.

Recently, the First Workshop on Scholarly Document Processing \cite{Chandrasekaran2020Overview}, Lay Summary Task\footnote{https://ornlcda.github.io/SDProc/index.html} (LaySumm 2020) first proposed the task of Lay Summary Generation. The task aims to generate summaries that are representative of the content, comprehensible and interesting to a lay audience. After checking the dataset that the task provides, we observe that lots of the sentences in lay summaries have corresponding sentences in original papers. Inspiring by this observation, we think that making binary sentence labels for extractive summarization and utilize them as extra supervision signals can help model generate better summaries. Therefore, we conduct BART \cite{lewis2019bart} encoder to make sentence representations and train extractive summarization together with abstractive summarization.

Experimental results show that leveraging sentence labels can improve the Lay summary generation performance. In the Laysumm 2020 competition, our model achieves 46.00\% Rouge1-F1 score. The code will be released on Github \footnote{https://github.com/TysonYu/Laysumm}.

\section{Related Work}
\paragraph{Text Summarization}
Text summarization aims to produce a condensed representation of input text that captures the core meaning of the original text. Recently, neural network-based approaches have reached remarkable performance for news articles summarization \cite{see2017get, liu2019text, zhang2019pegasus}. Comparing with news articles, scientific papers are typically longer and contain more complex concepts and technical terms. 

\paragraph{Scientific Paper Summarization}
Existing approaches for scientific paper summarization include extractive models that perform sentence selection \cite{qazvinian2013generating,cohan2017scientific,cohan2018scientific} and hybrid models that select the salient text first and then summarize it \cite{ subramanian2019extractive}. Besides, \citet{cohan2018discourse} built the first model for abstractive summarization of single, longer-form documents (e.g., research papers).

In order to train neural models for this task, several datasets have been introduced. The arXiv and PubMed datasets \cite{cohan2018discourse} were created using open access articles from the corresponding popular repositories. \citet{yasunaga2019scisummnet} developed and released the first large-scale manually-annotated corpus for scientific papers (on computational linguistics). 


\paragraph{Large Pre-trained Language Model}
Large pre-trained language models, such as BERT \cite{devlin2018bert}, UniLM \cite{dong2019unified} and BART \cite{lewis2019bart} have shown great performance on a variety of downstream tasks including summarization. For example, BART achieved state-of-the-art performance on CNN/DM \cite{hermann2015teaching} news summarization dataset. 


\begin{figure*}
    \centering
    {\includegraphics[width=0.82\linewidth]{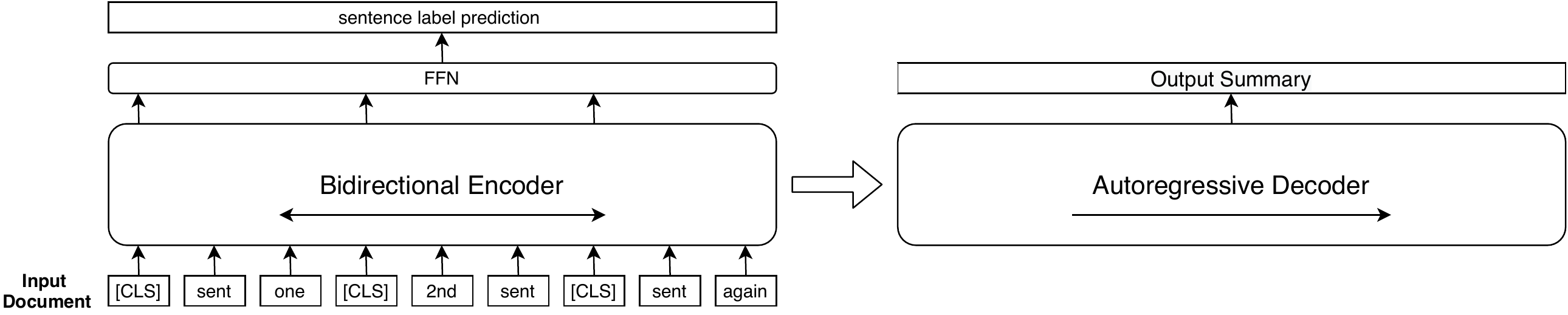}}
    \caption{Multi-label summarization model. The left part is based on a bidirectional encoder and the right part is an autoregressive decoder.}
    \label{fig:model}
\end{figure*}
\section{Datasets}
We use two datasets for this work, which are the dataset of CL-LaySumm 2020 and ScisummNet \cite{yasunaga2019scisummnet}. In this section, we introduce the details of them and the pre-processing method we used. 

\subsection{CL-LaySumm 2020 Dataset}
\label{CL-LaySumm 2020 Dataset}
The CL-LaySumm 2020 Dataset is released by the CL-LaySumm Shared Task that aims to produce lay summaries of scientific texts. A lay summary refers to a textual summary intended for a non-technical audience. There are 572 samples in the dataset for training and each sample contains a full-text paper with a lay summary. To test the summarization model, we need to generate lay summaries for 37 papers within 150 words.

Since the original papers are very long and the task requires us to generate relatively short summaries, it is crucial to extract important parts of papers first before feeding them to large pre-trained models. Given our own experience of how papers are written, we start with the assumption that the Abstract, Introduction and Conclusion are most likely to convey the topic and the contributions of the paper. So, we make different combinations of these three sections as input to our model. 

\subsection{ScisummNet Dataset}
The ScisummNet is the first large-scale, human-annotated Scisumm dataset. The dataset provides 1009 papers with their citation networks as well as their manual summaries. The gold summaries are written by annotators based on the abstract and selected citation sentences that also convey the contributions of papers. We take the abstract and annotators selected citation sentences as our models' input.

\subsection{Data Pre-processing}
As mentioned above, we first represent the document using the sentences in its Abstract, Introduction and Conclusion. Then we use two approaches to pre-process the text.

The first pre-processing approach is removing tags and outliers. The original text of the Laysumm dataset has lots of tags such as TITLE, SECTION and PARAGRAPH. We remove all different kinds of tags. Besides, some samples of the Laysumm dataset do not contain an Abstract or Introduction. We regard these samples as outliers and delete them while training the model. The total number of outliers is 23. Then, we truncate all input text to a max length of 1024 tokens due to the carrying capacity of the BART model.


\section{Methodology}
\begin{table*}
\centering
\resizebox{0.82\textwidth}{!}{%
\begin{tabular}{lrrrrrr}
\hline
\multicolumn{1}{l}{Model}               & Rouge1-F1 & Rouge1-Recall & Rouge2-F1 & Rouge2-Recall & RougeL-F1 & RougeL-Recall \\ \hline
BART (Abs)                         & 0.4350     & 0.4697        & 0.1807    & 0.1968        & 0.2722    & 0.2934        \\
BART (Abs+Intro)            & 0.4518    & 0.4923        & 0.1977    & 0.2135        & 0.2820     & 0.3061        \\
BART (Abs+Intro$_{all}$)            & 0.4443    & 0.4816        & 0.1991    & 0.2142        & 0.2825     & 0.3040        \\
BART (Abs+Intro+Con) & 0.4536    & \textbf{0.5171}        & 0.2016    & \textbf{0.2271}        & 0.2864    & \textbf{0.3243}        \\
BART (Data augmentation) & 0.4490	& 0.4887 & 0.1972 & 0.2136  & 0.2895 & 0.3139 \\
BART + Two-stage             & 0.4529    & 0.4882        & 0.2067    & 0.2224        & \textbf{0.2929}    & 0.3140         \\
BART + Multi-label                       & \textbf{0.4600}	 & 0.5013	 & \textbf{0.2070}	 & 0.2223	 & 0.2876	 & 0.3104        \\ \hline
\end{tabular}
}
\caption{Our results on CL-LaySumm 2020 shared task.}
\label{tab: results}
\end{table*}

\subsection{Baseline}
We use BART, a denoising autoencoder for pretraining sequence-to-sequence models \cite{lewis2019bart} as our baseline. 

BART is based on the standard Transformer model \cite{vaswani2017attention}, which can be regarded as generalizing BERT (due to the bidirectional encoder), GPT (with the left-to-right decoder). It is pre-trained on the same corpus as RoBERTa \cite{liu2019roberta} with two tasks:  text infilling and sentence permutation. For text infilling, 30\% of tokens in each document are masked and the model is trained to recover them at the output. For the sentence permutation, all sentences are permuted as input and the model is supposed to generate the output sentences with the correct order.

BART obtains great performance on the summarization task. We use the BART fine-tuned on CNN/DailyMail dataset \cite{hermann2015teaching} to initialize our model. 

\subsection{Multi-Label Summarization Model}
\label{Multi-Label Summarization Model}
There are two canonical strategies for summarization: extractive summarization, which concatenates sentences into the summary and abstractive summarization, which generate novel sentences for the summary. Inspired by the observation that lots of the sentences in human written lay summaries have corresponding sentences in original papers, we use an unsupervised approach to convert the abstractive summaries to extractive labels and train abstractive summarization together with extractive summarization.

To make the ground truth sentence-level binary labels for extractive summarization, which we call ORACLE, we use a greedy algorithm introduced by \cite{nallapati2017summarunner}. The approach is based on the idea that the selected sentences from the input should be the ones that maximize the Rouge score \cite{lin2003automatic} with the respect gold summary.

The architecture of our model is shown in Figure \ref{fig:model}, which follows the BART model's structure. The input document is fed into the bidirectional encoder, then the contextual embeddings of the $i^{th}$ [CLS] symbol are used as the sentence representations. After a feedforward neural network, these sentence representations produce a binary distribution about whether they belong to the extractive summary. As for the abstractive summary, it is generated by the autoregressive decoder. The overall loss $L$ is calculated by $L = w_{e}L_{e} + L_{a}$. Here $L_{e}$ and $L_{a}$ refer to the Cross-Entropy loss of extractive and abstractive summary respectively.

\subsection{Data Augmentation}
Data augmentation has been an effective technique to create new training instances when the training data is not enough, as demonstrated in computer vision as well as for many NLP tasks \cite{chen2017reading, yang2019data, yuan2017machine}.


Existing data augmentation approaches in NLP tasks can be categorized into retrieval-based methods \cite{chen2017reading, yang2019data} and generation-based methods \cite{yuan2017machine, buck2017ask}. However, none of these suits our situation, since external sources or auxiliary training data are still required. So we adopted a similar method from \cite{nema2017diversity}. A pre-defined vocabulary of 24,822 words was used where each word had been associated with a synonym. Then for each training instance, certain ratios (in our case, 1/9) in each document were randomly selected (except stop words and numerical values) and then replaced with their synonyms found in the vocabulary. If a selected word was not found in the vocabulary, it was added there with the most similar word found based on cosine similarity in the GloVe \cite{pennington2014glove} vocabulary. For each training instance, this process is repeated 9 times to create 9 new documents. But the same summary of the original instance was used in the newly generated instances. 

\subsection{Two-Stage Fine-tuning}
\label{two stage fine-tuning}
To make use of the ScisummNet dataset, we conduct a two-stage fine-tuning method. In the first stage, we fine-tune the pre-trained BART model on the ScisummNet dataset. We use the Abstract and annotators selected citation sentences as the input and the gold summary as the output. The model is fine-tuned with 20000 iterations before saved. As for the second stage, we use the same settings as we directly fine-tune on the CL-LaySumm 2020 dataset.

\section{Experiments}
During the training phase, we randomly select 90\% of the CL-LaySumm 2020 Dataset for training and 10\% for validation. If a data sample doesn't contain an Abstract or Introduction, we don't include it in training or validation. To find the optimal architecture for this task within the models we have, we set up seven different experiments. 

\textit{BART (Abs)}: We only use the Abstract as the input to the BART model. 

\textit{BART (Abs+Intro)}: We use the Abstract and the first paragraph of the Introduction as the input to the BART model.

\textit{BART (Abs+Intro$_{all}$)}: We use the Abstract and the whole Introduction as the input to the BART model.

\textit{BART (Abs+Intro+Con)}: We use the Abstract, the first paragraph of the Introduction, and the Conclusion (if the paper has) as the input to the BART model. 

\textit{BART (Data augmentation)}: We use the same data as BART (Abs+Intro+Con). For each training sample, we create 9 new input documents by synonym data augmentation. 

\textit{BART + Two-stage}: We use the same data as BART (Abs+Intro+Con) to the BART model. The two-stage fine-tuning method is introduced in Section \ref{two stage fine-tuning}

\textit{BART + Multi-label}: We use the same data as BART (Abs+Intro+Con). In addition, for each sentence in the input, we add [CLS] token at the beginning.

As for the hyperparameters, we use a dynamic learning rate, warm up 1000 iterations, and decay afterward. We set the batch size to 1 because of the limitation of GPU memory. The gradient will accumulate every ten iterations and we train all models for 6000 iterations on 1 GPU (GTX 1080 Ti). We save the best model that has the highest Rouge1-F1 score based on the validation set. For the BART model, we use the implementation from the huggingface\footnote{https://github.com/huggingface/transformers}. We use the BART large model pre-trained on CNN/DailyMail dataset.

\section{Result Analysis}
The results are shown in Table \ref{tab: results} and we analyze them from three aspects. Besides, we also generate a Lay Summary of our paper, which is presented in the appendix ~\ref{our own paper}.

\paragraph{Different inputs to the model.}
The experiment results of BART (Abs), BART (Abs+Intro), and BART (Abs+Intro+Con) show by adding the Introduction and Conclusion to the input, the models' performance improves consistently. However, comparing with the results from BART (Abs+Intro) and  BART (Abs+Intro$_{all}$), using the whole Introduction rather than the first paragraph of the Introduction decreases the performance on Rouge1 score. We think it is because the CL-LaySumm 2020 task requires to make a relatively short summary, less than 150 words. If the input is too long, it makes the model harder to summarize because longer input contains more noisy data. Since the CL-LaySumm 2020 dataset is also small, the model doesn't have enough samples to learn the task.

\paragraph{Two-stage fine-tuning and Data Augmentation.}
The experimental results show that two-stage fine-tuning doesn't help to improve the model's performance. After checking the details of ScisummNet, we find the corpus comes from ACL Anthology Network (AAN) \cite{radev2013acl}, which means all data relates to computational linguistics. In contrast, the CL-LaySumm 2020 dataset use papers from a variety of domains including biology and medicine. The Statistical differences between these two datasets make the model hard to learn prior knowledge that can be utilized in CL-LaySumm 2020 task.

As for the Data Augmentation, the model performance also doesn't increase as we expected, which contradicts the results from the original paper \cite{nema2017diversity}. However, the same method also fails in \cite{laskar2020query}, which also adopted a large pre-trained model as a start-point for fine-tuning. So we think the possible reason might be that large pre-trained models are less robust to noisy input. Our synonyms replacement method is too simple as well as unsupervised. On one hand, it can increase the vocabulary diversity of the training data without changing the semantic meaning a lot, but on the other hand, the quality especially the grammar of the generated instances can not be guaranteed to be correct. Thus, some noise might be introduced and decreases the model performance when we augment the data.


\paragraph{Multi-label summarization.}
Comparing with BART (Abs+Intro+Con) and BART + Multi-label models, we find that with multi labels, the Rouge1-F1 score is better but the Recall score is lower, which means that the precision increase a lot. We think that with the extra supervision of sentence labels, the model can learn a better sentence understanding. As a result, the model is able to extract important content from the input which helps upper the F1 and Precision scores.

\section{Conclusion}
In this paper, we showcased how different inputs, data augmentation, training strategy, and sentence labels influence the lay summarization task. We introduce a new method to utilize sentence labels as another supervision signal while training BART based model. Experimental results show our models can generate better summaries evaluated by the Rouge1-F1 score.

\normalem
\bibliographystyle{acl_natbib}
\bibliography{emnlp2020}

\clearpage
\appendix
\setcounter{table}{0} 
\setcounter{figure}{0}
\renewcommand{\thetable}{\Alph{section}\arabic{table}}
\renewcommand\thefigure{\Alph{section}\arabic{figure}} 


\section{Case Study}
\label{our own paper}
\subsection{The Lay Summary of this Paper}
In the CL-LaySumm 2020 shared task, our model achieves 46.00\% Rouge1-F1 score. In this paper, we build a lay summary generation system based on the BART model. We leverage sentence labels as extra supervision signals to improve the performance of lay summarization. Experimental results show that leveraging sentence labels can improve the Lay summary generation performance. The code will be released on Github.
\subsection{Observation}
The summary above is generated by our own system with Abstract, Introduction and Conclusion from this paper. Although many sentences are copied from the original text, they are well organized and coherent. Besides, the content of the summary also conveys the topic and the contribution of this paper. In conclusion, our system can produce accurate and readable summaries.

\end{document}